\documentclass[conf]{IEEEtran}
\usepackage{multicol}
\usepackage{color}

\usepackage{amssymb,amsmath,amsfonts}
\usepackage{amsthm}
\usepackage{algorithm}
\usepackage{subfigure}
\usepackage{lipsum}
\usepackage[noend]{algpseudocode}
\usepackage{color}
\usepackage{todonotes}
\usepackage{url}
\usepackage{chngcntr}
\usepackage{thmtools}
\usepackage{cite}
\usepackage{epstopdf}
\usepackage{mathtools}
\newcommand{\ie}[1]{\textit{i.e.,}}

\makeatletter
\def\BState{\State\hskip-\ALG@thistlm}
\makeatother
\newcommand{\eg}{\textit{e.g.}}
\DeclareMathOperator*{\argmax}{arg\,max}

\newcommand{\revone}[1]{{\color{black}{#1}}}

\begin{document}
\title{\LARGE \bf
Multi-Fidelity Reinforcement Learning with Gaussian Processes
}

\author{Varun Suryan,$^{1}$  Nahush Gondhalekar,$^{2}$ and Pratap Tokekar$^{3}$
\thanks{*This work has been funded by the Center for Unmanned Aircraft Systems (C-UAS), a National Science Foundation-sponsored industry/university cooperative research center (I/UCRC) under NSF Award No. IIP-1161036 along with significant contributions from C-UAS industry members. This work was performed when the authors were with the Department of Electrical \& Computer Engineering, Virginia Tech, U.S.A.}
\thanks{$^{1}$Varun Suryan is with the Department of Computer Science, University of Maryland College Park, U.S.A.
        {\tt\small suryan@umd.edu}}%
\thanks{$^{2}$Nahush Gondhalekar is with Qualcomm, Inc.
        {\tt\small nahushg@vt.edu}}%
\thanks{$^{3}$Pratap Tokekar is with the Department of Computer Science, University of Maryland College Park, U.S.A.
        {\tt\small tokekar@umd.edu}}%
}

\maketitle

\begin{abstract}
We study the problem of Reinforcement Learning (RL) using as few real-world samples as possible. A naive application of RL can be inefficient in large and continuous state spaces. We present two versions of Multi-Fidelity Reinforcement Learning (MFRL), model-based and model-free, that leverage Gaussian Processes (GPs) to learn the optimal policy in a real-world environment. In the MFRL framework, an agent uses multiple simulators of the real environment to perform actions. With increasing fidelity in a simulator chain, the number of samples used in successively higher simulators can be reduced. \revone{By incorporating GPs in the MFRL framework, we empirically observe up to $40\%$ reduction in the number of samples for model-based RL and $60\%$ reduction for the model-free version.} We examine the performance of our algorithms through simulations and through real-world experiments for navigation with a ground robot.
\end{abstract}

\IEEEpeerreviewmaketitle

\section{Introduction}
\IEEEPARstart{R}{ecently}, there has been a significant development in reinforcement learning (RL) for robotics. A major limitation of using RL for planning with robots is the need to obtain a large number of training samples. Obtaining a large number of real-world training samples can be expensive and potentially dangerous. In particular, obtaining exploratory samples--which are crucial to learning optimal policies--may require the robot to collide or fail, which is undesirable. Motivated by these scenarios, we focus on minimizing the number of real-world samples required for learning optimal policies.


\revone{One way to reduce the number of real-world samples is to leverage simulators}~\cite{chebotar2018closing}. Collecting learning samples in a robot simulator is often inexpensive and fast. \revone{One can use a simulator to learn an initial policy which is then transferred to the real-world --- a technique  usually referred to as sim2real~\cite{chebotar2018closing}.} This lets the robot to avoid learning from scratch in the real world and hence, \revone{reducing the number of physical interactions} required. However, this comes with a trade-off. While collecting learning samples in simulators is inexpensive, \revone{they often fail to capture the real-world environments perfectly, a phenomenon called as the reality gap. While simulators with increasing fidelity with respect to the real-world are being developed, one would expect there to always remain some reality gap.} 

\begin{figure}[htp]
	\centering
	\includegraphics[width=1\columnwidth]{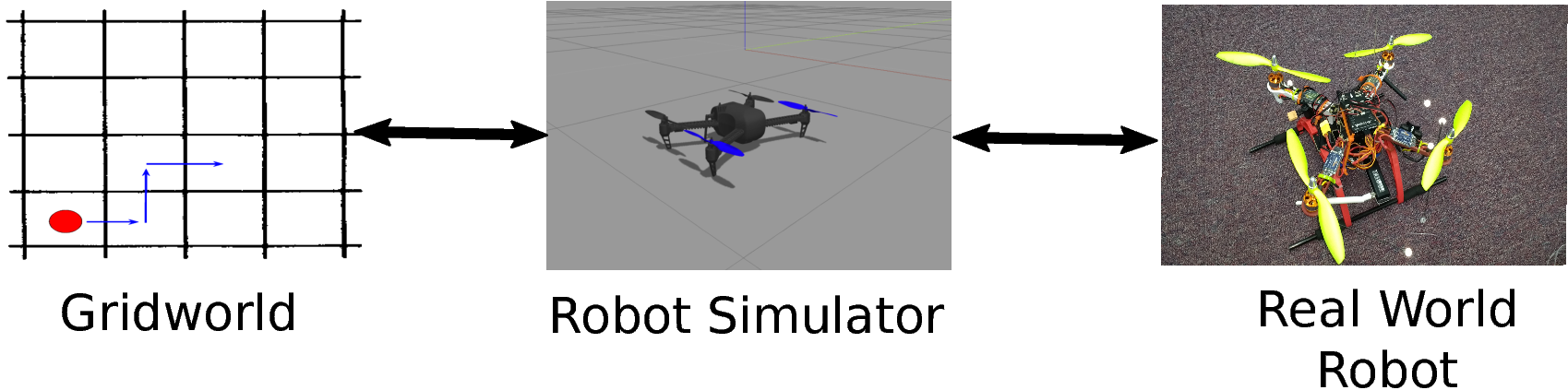}
	\caption {MFRL framework: The first simulator captures only grid-world movements of a point robot while the second simulator has more fidelity modeling the physics as well. Control can switch back and forth between simulators and real environment which is the third simulator in the chain.}
   \label{fig:mfrl_architecture}
\end{figure}
\revone{In this paper, we leverage the concept of} Multi-Fidelity Reinforcement Learning (MFRL) algorithm~\cite{cutler2014reinforcement} that uses multiple simulators with varying fidelity levels to minimize the number of real-world (\ie, highest-fidelity simulator) samples. The simulators, denoted by $\Sigma_1,\ldots,\Sigma_d$, have increasing levels of fidelity \revone{with respect to} the real environment. For example, $\Sigma_1$ can be a simple simulator that models only the robot kinematics, $\Sigma_2$ can model the dynamics as well as kinematics, and the highest fidelity simulator can be the real world (Figure~\ref{fig:mfrl_architecture}).

MFRL differs from transfer learning~\cite{taylor2007transfer}, where a transfer of parameters is allowed only in one direction. The MFRL algorithm starts in $\Sigma_1$. Once it learns a sufficiently good policy in $\Sigma_1$, it switches to a higher fidelity simulator. If it observes that the policy learned in the lower fidelity simulator is no longer optimal in the higher fidelity simulator, it switches back to the lower fidelity simulator. \cite{cutler2014reinforcement} showed that the resulting algorithm has polynomial sample complexity and minimizes the number of samples required for the highest fidelity simulator. 

The original MFRL algorithm learns the transition and reward functions at each level. The reward and transition for each state-action pair are learned independently of others. While this is reasonable for general agents, when planning for physically-grounded robots, we can exploit the spatial correlation between neighboring state-action pairs to speed up the learning. 

Our main contribution is to leverage the GP regression as a function approximator to speed up learning in the MFRL framework. GPs can predict the learned function value for any query point, and not just for a discretized state-action pair. Furthermore, GPs can exploit the correlation between nearby state-action values by an appropriate choice of a kernel. GPs have been extensively used to obtain optimal policies in simulation-aided reinforcement learning~\cite{cutler2015efficient}. We take this further by using GPs in the MFRL setting.

\revone{Other function approximators have been used in RL previously. We choose to use GPs since they require fewer samples to learn a function when a good prior  exists~\cite{neal2012bayesian}. The priors can be imposed, in part, by using appropriate kernels which make GPs flexible. A major limitation of learning with GPs is their computational complexity, which grows cubically with respect to the number of training samples. However, this issue can be mitigated by using sparse approximations for GPs~\cite{quinonero2005unifying}.}
In MFRL, the state-space of $\Sigma_i$ is a subset of the state space of $\Sigma_j$ for all $j>i$. Therefore, when the MFRL algorithm switches from $\Sigma_i$ to $\Sigma_{i+1}$ it already has an estimate for the transition function and $Q$--values for states in $\Sigma_{i+1}$. Hence, GPs are particularly suited for MFRL, which we verify through our simulation results.

Our main contributions in this paper include introducing:
\begin{enumerate}
\item a model-based MFRL algorithm, GP-VI-MFRL, which estimates the transition function and subsequently calculates the optimal policy using Value Iteration (VI); and
\item a model-free MFRL algorithm, GPQ-MFRL, which directly estimates the optimal $Q$-values and subsequently the optimal policy.
\end{enumerate}

We verify the performance of the algorithms presented through simulations as well as experiments with a ground robot. Our empirical evaluation shows that the GP-based MFRL algorithms learn the optimal policy faster than the original MFRL algorithm using even fewer real-world samples. 

The rest of the paper is organized as follows. In the next section, we present a survey of related work followed by the background on RL and GPs in Section~\ref{sec:background}. Section~\ref{sec:alg} presents our algorithms (GP-VI-MFRL and GPQ-MFRL). We present the experimental results in Section~\ref{section:results} along with comparisons with other MFRL and non-MFRL techniques. We conclude with a discussion of future work.

\section{Related Work}\label{related:work}
Multi-fidelity methods are prominently used in various engineering applications. These methods are used to construct a reliable model of a phenomenon when obtaining direct observations of that phenomenon are expensive. The assumption is that we have access to cheaply-obtained, but possibly less accurate observations from an approximation of that phenomenon. Multi-fidelity methods can be used to combine those observations with expensive but accurate observations to construct a model of the underlying phenomenon~\cite{cutler2015efficient}. For example, learning the dynamics of a robot using real-world observations may cause wear and tear of the hardware~\cite{kober2013reinforcement}. Instead, one can obtain observations from a simulator that uses a, perhaps crude, approximation of the true robot dynamics~\cite{tan2018sim}. 

\revone{Let} $f: \mathcal{X} \rightarrow \mathcal{Y}$ \revone{denote a function that maps the} input $\mathbf{x}\in\mathcal{X}\subset\mathbb{R}^d$ to an output $\mathbf{y}\in\mathcal{Y}\subset\mathbb{R}^{d^{'}}$, where $d, d^{'}\in \mathbb{N}$. Multiple approximations are available that estimate the same output with varying accuracy and costs. 
More generally, $K$ different fidelity approximations, $f^{(1)}, \ldots, f^{(K)}$, are available representing the relationship between the input and the output, $f^{(i)}:\mathcal{X}\rightarrow\mathcal{Y}, i =1,\ldots, K$.
Obtaining observations from the $i^{th}$ approximation incurs cost $c^{(i)}$, and typically $c^{(i)}<c^{(j)}$ for $i<j$.

There has been a significant surge in multi-fidelity methods research following the seminal work on autoregressive schemes~\cite{kennedy2000predicting}. They used GPs to explore ways in which runs from several levels of a computer code can be used to make inference about the output of the complex computer code. A complex code approximates the reality better, but in extreme cases, a single run of a complex code may take many days. GP framework is a natural candidate to estimate a phenomenon by combining data from various fidelity approximations~\cite{perdikaris2017nonlinear, damianou2015deep}.

Multi-fidelity methods have been widely used in RL applications to automatically optimize the parameters of control policies based on data from simulations and experiments. 
\revone{Cutler et al.}~\cite{cutler2014reinforcement} introduced a framework which specifies the rules on when to collect observations from various fidelity simulators. \revone{Marco et al.}~\cite{marco2017virtual} introduced a \revone{Bayesian Optimization} algorithm which uses entropy~\cite{hennig2012entropy} as a metric to decide which simulator to collect the observations from. They used their algorithm for a cart-pole setup with a Simulink model as the simulator.

\revone{Two closely related techniques addressing sim2real are domain randomization and domain adaptation. In both cases, the simulators can be controlled via a set of parameters. The underlying hypothesis is that some unknown set of parameters closely match the real-world conditions. In domain randomization, the simulator parameters are randomly sampled and the agent is trained across all the parameter values. In domain adaptation, the parameter values are updated during learning. However, it is important to note that the goal in these methods is to reduce the reality gap by using a parameterized simulator~\cite{tobin2017domain,chebotar2018closing}. This restricts the use of such approaches to scenarios where altering the parameters for a simulator itself is trivial. Further, only one type of simulator is used. In contrast, MFRL techniques leverages multiple simulators with varying fidelity levels as well as cost to operate. Furthermore, in MFRL the policy learned in the highest fidelity simulator (real-world) uses data from the same environment, unlike sim2real. As such, these approaches are beneficial when there is a significant reality gap where maneuvers learned in simulators may not translate to the real world.}

Our work is inspired by the work of \cite{cutler2014reinforcement} which allows for bidirectional transfer of information between simulators. However, they used a tabular representation of the values function. We introduce two algorithms in a similar spirit which can decide which simulator to collect the observations from. We hypothesize that by using a GP with MFRL framework will lead to further improvements in \revone{the number of samples required from} the real world. 

\section{Background}\label{sec:background}
\subsection{Reinforcement Learning}
RL problems can be formulated as a Markov Decision Process (MDP): $\mathcal{M}=\langle \mathcal{S}, \mathcal{A}, \mathcal{P}, \mathcal{R}, \gamma\rangle;$ with state space $\mathcal{S}$; action space $\mathcal{A}$; transition function $\mathcal{P}(s_t, a_t, s_{t+1}) \mapsto [0, 1]$; reward function $\mathcal{R}(s_t, a_t)\mapsto \mathbb{R}$ and discount factor $\gamma\in[0,1)$. A
policy $\pi : \mathcal{S} \rightarrow \mathcal{A}$ maps states to actions. Together with the initial state $s_0$, a policy forms a trajectory $\zeta = \lbrace[s_0, a_0, r_0], [s_1, a_1, r_1], \ldots\rbrace$ where $a_t = \pi(s_t)$. $r_t$ and $s_{t+1}$ are sampled from the reward and transition functions, respectively. 

We consider a scenario where the goal is to maximize the infinite horizon discounted reward starting from a state $s_0$. The value function for a state $s_0$ is defined as $\mathcal{V}^{\pi}(s_0)=\mathbb{E}[ \sum_{t=0}^{t=\infty}\gamma^tr_{t}(s_t, a_t)|a_t = \pi(s_t)]$. The state-action value function or $Q$-value of each state-action pair under policy $\pi$ is defined as $Q^{\pi}(s, a) = \mathbb{E}[\sum_{t=0}^{t=\infty}\gamma^tr_{t+1}(s_{t+1},a_{t+1})|s_0=s,a_0=a]$ which is the expected sum of discounted rewards obtained starting from state $s$, taking action $a$ and following $\pi$ thereafter. The optimal $Q$-value function $Q^{*}$ for a state-action pair $(s, a)$ satisfies $Q^{*}(s, a)=\textnormal{max}_{\pi}Q^{\pi}(s, a)=\mathcal{V}^{*}(s)$ and can be written recursively as,
\begin{align}\label{eq:bellman}
Q^{*}(s_t,a_t)=\mathbb{E}_{s_{t+1}}\big[r(s_t, a_t)+\gamma\mathcal{V}^{*}(s_{t+1})\big].
\end{align}

Our objective is to find the optimal policy $\pi^{*}(s)=\textnormal{argmax}_aQ^{*}(s, a)$ when $\mathcal{R}$ and $\mathcal{P}$ are not known to the agent. In model-based approaches, the agent learns $\mathcal{R}$ and $\mathcal{P}$ first and then finds an optimal policy by calculating optimal $Q$-values from Equation~\eqref{eq:bellman}. The most commonly used model-based approach is VI~\cite{ECML10-jung, brafman2002r}. We can also directly estimate the optimal $Q$-values, often known as model-free approaches~\cite{grande2014sample} or directly calculate the optimal policy, often known as policy-gradient approaches~\cite{sutton1998reinforcement}. The most commonly used model-free algorithm is $Q$-learning. 
For the GP-VI-MFRL implementation, we use GPs to estimate transition function and value iteration to calculate the optimal policy. For our GPQ-MFRL implementation, we use $Q$-learning to perform the policy update using GP regression.

In this work, two versions (model-based and model-free) of GP based MFRL are introduced by keeping the fact in mind that while model-based algorithms are generally more sample efficient in comparison to model-free algorithms but they are not memory efficient~\cite{brafman2002r}. Hence, depending on the application, model-free approach (GPQ-MFRL) may be a suitable alternative to more sample efficient model-based GP-VI-MFRL.  

\subsection{Gaussian Processes}
GPs are Bayesian non-parametric function approximators. GPs can be defined as a collection of infinitely many random variables, any finite subset $\mathbf{X}=\lbrace \mathbf{x}_1, \ldots,\mathbf{x}_k \rbrace$ of which is jointly Gaussian with mean vector $\mathbf{m}\in\mathbb{R}^k$ and covariance matrix $\mathbf{K}\in \mathbb{R}^{k\times k}$~\cite{rasmussen2006gaussian}.

Let $\mathbf{X}=\lbrace \mathbf{x}_1, \ldots,\mathbf{x}_k \rbrace$ denote the set of the training inputs. Let $\mathbf{y}=\lbrace y_1, \ldots,y_k \rbrace$ denote the corresponding training outputs. GPs can be used to predict the output value at a new test point, $\mathbf{x}$, conditioned on the training data. Predicted output value at $\mathbf{x}$ is normally distributed with mean $\hat{\mu} (\mathbf{x})$ and variance $\hat{\sigma}^2(\mathbf{x})$ given by,
\begin{align}
&\hat{\mu} (\mathbf{x}) = \mu(\mathbf{x}) + \mathbf{k}(\mathbf{x}, \mathbf{X})\Big[\mathbf{K}(\mathbf{X},\mathbf{X})+\omega^2\mathbf{I}\Big]^{-1} \mathbf{y} \label{eq:predMean}, \\
&\hat{\sigma}^2(\mathbf{x}) = k(\mathbf{x},\mathbf{x}) - \mathbf{k}(\mathbf{x}, \mathbf{X})\Big[\mathbf{K}(\mathbf{X},\mathbf{X})+\omega^2\mathbf{I}\Big]^{-1} \mathbf{k}(\mathbf{X}, \mathbf{x}), \label{eq:predVar}
\end{align}
where $\mathbf{K}(\mathbf{X},\mathbf{X})$ is the kernel. The entry $\mathbf{K}_{\mathbf{x}_l,\mathbf{x}_m}$ gives the covariance between two inputs $\mathbf{x}_l$ and $\mathbf{x}_m$. $\mu(\mathbf{x})$ in Equation~\eqref{eq:predMean} is the prior mean of output value at $\mathbf{x}$. 

We use a zero-mean prior and a squared-exponential kernel where $\mathbf{K}_{\mathbf{x}_l,\mathbf{x}_m}$ is given by,
\begin{align}
\mathbf{K}_{\mathbf{x}_l,\mathbf{x}_m} = \sigma^2\exp\left(-\frac{1}{2}\sum_{d=1}^{d=D}\left(\frac{\mathbf{x}_{dl}-\mathbf{x}_{dm}}{l_d}\right)^2\right)+\omega^2,
\end{align}
$\sigma^2,~l_d~\textnormal{and}~\omega^2$ are hyperparameters that can be either set by the user or learned online through the training data. \revone{$D$ is the dimension of the training inputs.}

In the GPQ-MFRL algorithm, we use GPs to learn $Q$--values. GPs are proved to be consistent function approximators in RL with convergence guarantees~\cite{rasmussen2006gaussian}. A set of state-action pairs is the input to GP and $Q$--values are the output/observation values to be predicted. In GP-VI-MFRL algorithm, we use GPs to learn the transition function. The input to the GPs is a set of observed state-action pairs and we predict the next state as output for a newly observed state-action pair.

\section{Algorithm Description}\label{sec:alg}
\begin{figure}[htp]
	\centering
	\includegraphics[width=0.6\columnwidth]{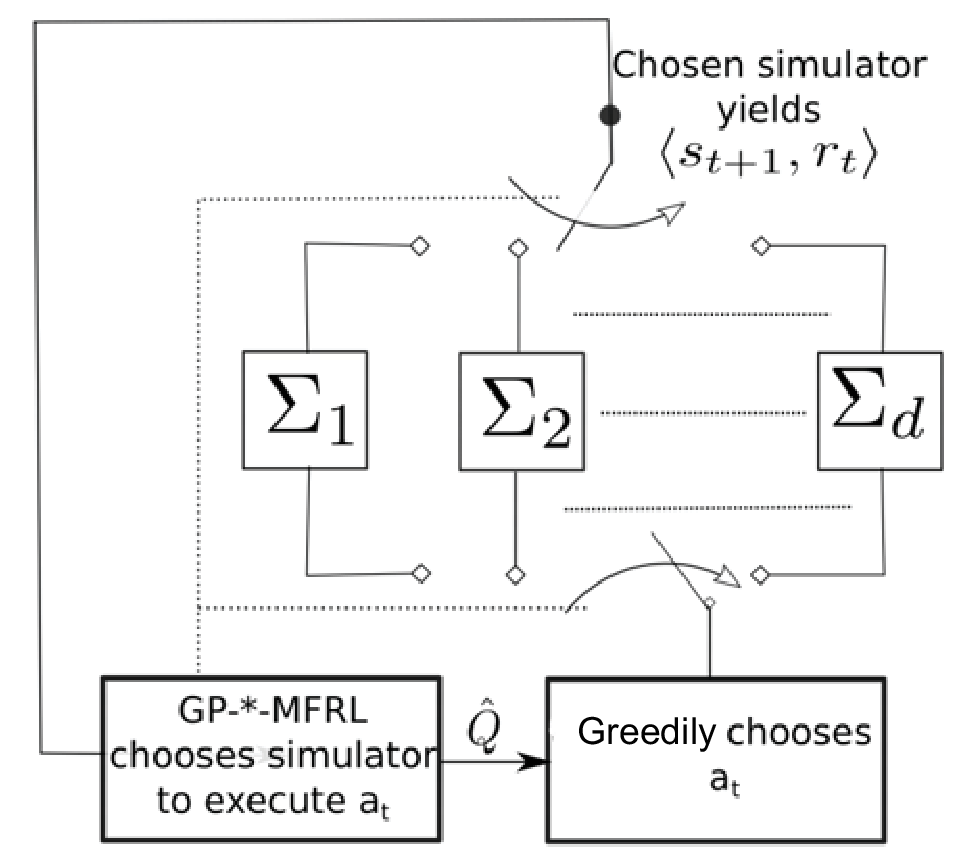}
	\caption {The simulators are represented by $\Sigma_1, \Sigma_2, \ldots, \Sigma_d$. The algorithms decide the simulator in which the current action is to be executed by the agent. Also, the action values in the chosen simulator are updated and used to select the best action using the information from higher as well as lower simulators.}
   \label{fig:gp_mfrl_system}
\end{figure}

In this section, we first describe both versions of our algorithm.
We compare the proposed algorithms with baseline strategies through simulations. A flow chart of our algorithms is shown in Figure~\ref{fig:gp_mfrl_system}. 
We make the following assumptions for both algorithms.
\begin{enumerate}
\item The reward function is known to the agent. We make this assumption for the ease of exposition. In general, one can use GPs to estimate the reward function as well. \revone{This assumption is required only for GP-VI-MFRL algorithm.}
\item State-space in simulator $\Sigma_{i-1}$ is a subset of the state-space in simulator $\Sigma_i$. The many-to-one mapping $\rho_i$ maps states from simulator $\Sigma_i$ to states in simulator $\Sigma_{i-1}$. We give an example of such mapping in subsequent sections. Let $\rho_i^{-1}$ denote the respective inverse mapping (which can be one-to-many) from states in $\Sigma_{i-1}$ to states in $\Sigma_{i}$. \revone{The action space is discrete and same for all the simulators and real world.} 
\end{enumerate}

\subsection{GP-VI-MFRL Algorithm}
\begin{algorithm}[!htp]
\caption{GP-VI-MFRL Algorithm}\label{mbGP:MFRL}
\begin{algorithmic}[1]
\Procedure{}{}
\BState \textbf{Input:} confidence parameters $\sigma_{th}$ and $\sigma^{sum}_{th}$; simulator chain $\langle \text{\boldmath$\Sigma$} 
,~\textnormal{fidelity parameters}~\text{\boldmath$\beta$},~\textnormal{state mappings}~\text{\boldmath$\rho$} \rangle;~\mathcal{L}$. 
\BState \textbf{Initialize:} Transition functions $\mathcal{P}^a_{ss'}(i)$ and $\mathcal{D}_i$ for $i\in\lbrace 1, \ldots, d\rbrace$; $change = \textsc{False}$. 
\BState \textbf{Initialize:} $i~\gets~1;~\hat{Q}_i~\gets~$ \textsc{Planner} $\left(\mathcal{P}^a_{ss'}(i)\right)$. 
\BState \textbf{while} terminal condition is not met
\State \hspace{.3cm} $a_t~\gets~\argmax_a \hat{Q}_i(s, a)$ 
\State \hspace{.3cm} \textbf{if} $\sigma_i(s_t, a_t)\leq \sigma_{th}$: $change=\textsc{True}$
\State \hspace{.3cm} \textbf{if} $\sigma\left(\rho_i(s_t), a_t\right) \geq \sigma_{th}$ and $change$ and $i>\textnormal{1}$
\State \hspace{1cm} $s_t~\gets~\rho_i(s_t),~i~\gets~{i - 1}$, continue 
\State \hspace{.30cm} $\langle s_{t+1}, r_{t+1}\rangle~\gets~\textnormal{execute}~a_t~\textnormal{in $\Sigma_i$}$
\State \hspace{.3cm} $\textnormal{append}~\langle s_t, a_t, s_{t+1}, r_t \rangle~\textnormal{ to $\mathcal{D}_i$}$
\State \hspace{.3cm} $\mathcal{P}^a_{ss'}(i)~\gets~\textnormal{update $\textnormal{GP}_i$ using}~\textnormal{$\mathcal{D}_i$}$
\State \hspace{.3cm} $\hat{Q}_i~\gets~\textnormal{call \textsc{Planner} with input}~\mathcal{P}_{ss'}^a(i)$
\State \hspace{.3cm} $t~\gets~{t+1}$ 
\State \hspace{.3cm} \textbf{if} $\sum_{j = t - \mathcal{L}}^{j=t-1}\sigma_i(s_j, a_j) \leq \sigma_{th}^{sum}~\textnormal{and}~t>\mathcal{L}~\textnormal{and}~i<d $ 
\State \hspace{.3cm} \ \ \ \ \ \ \ $s_t~\gets~\rho_{i+1}^{-1}\left(s_t\right),~i~\gets~{i+1};$ 
\State \hspace{.3cm} \ \ \ \ \ \ \ $change=\textsc{False}$ 
\BState \textbf{end procedure}
\EndProcedure  \\
\Procedure{Planner$\left(\mathcal{P}_{ss'}^a(i)\right)$}{}
\BState \textbf{Initialize:} $Q(s, a)=0~\textnormal{for each}~(s,a),~\Delta = \infty$
\BState \textbf{while} $\Delta > 0.1$
\State $\Delta~\gets~0$
\State \textbf{for} every $(s,a)$
\State \ \ \ \   $temp~\gets~Q(s, a),~Q(s, a) = \hat{Q}_{i-1}\left(\rho_i(s), a\right) + \beta_i$
\State \ \ \   \textbf{for} $k\in\lbrace i, \ldots, d\rbrace$ 
\State \ \ \ \ \ \ \ \ \  $s_k = \rho^{-1}_{k}\ldots\rho^{-1}_{i+2}\rho^{-1}_{i+1}(s)$ 
\State \ \ \ \ \ \ \ \ \ \textbf{if} $\sigma_k(s_k, a)\leq\sigma^{th}$: $\mathcal{P}_{ss'}^a(i) = \mathcal{P}_{ss'}^a (k)$ 
\State \ \ \ \ \ \  $Q(s, a)~\gets~\sum_{a}\sum_{s'} \mathcal{P}_{ss'}^a[\mathcal{R}_{ss'}^a+\gamma\textnormal{max}_a Q(s', a)]$
\State \ \ \  \ $\Delta \leftarrow \textnormal{max}(\Delta, \  |temp - Q(s, a)|)$
\BState return $Q(s, a)$
\BState \textbf{end procedure}
\EndProcedure
\end{algorithmic}
\end{algorithm}

GP-VI-MFRL consists of a model learner and a planner. The model learner learns the transition functions using GP-regression. We use VI~\cite{sutton1998reinforcement} as our planner to calculate the optimal policy with learned transition functions. Let $\textbf{s}_{t+1} = f(\textbf{s}_t, a_t)$ be the (unknown) transition function that must be learned. We observe transitions: $\mathcal{D} = \{\langle\textbf{s}_t, a_t, \textbf{s}_{t+1}\rangle\}$. Our goal is to learn an estimate $\hat{f}(\textbf{s}, a)$ of $f(\textbf{s}, a)$. We can then use this estimated $\hat{f}$ for unvisited state-action pairs (in place of $f$) during VI to learn the optimal policy. 
For a given state-action pair $(s,a)$, the estimated transition function is defined by a normal distribution with the mean and variance given by Equations~\eqref{eq:predMean} and~\eqref{eq:predVar}. Algorithm~\ref{mbGP:MFRL} gives the details of the proposed framework.

Before executing an action, the agent checks (Line 8) if it has a sufficiently accurate estimate of the transition function for the current state-action pair in the previous simulator, $\Sigma_{i-1}$. Specifically, we check if the variance of the current state-action pair in previous simulator is less than $\sigma_{th}$. If not, and if the transition model in the current environment has changed, it switches to $\Sigma_{i-1}$ and executes the action in the potentially less expensive environment. The agent lands in the state $\rho_i(s)$ in lower fidelity simulator. 

We also keep track of the variance of the $\mathcal{L}$ most recently visited state-action pairs in the current simulator. If the running sum of the variances is below a threshold (Line 15), this suggests that the robot is confident about its actions in the current simulator and can advance to the next one. In the original work~\cite{cutler2014reinforcement}, the agent switches to the higher fidelity simulator after a certain number of known state-action pairs were encountered. In our implementation (Line 7), the model of current environment changes if the posterior variance for a state-action pair drops below a threshold value (\ie,  agent has a sufficiently accurate estimate of the transitions from that state). Lines 10--13 describe the main body of the algorithm, where the agent executes the greedily chosen action and records the observed transition in $\mathcal{D}_i$. The GP model for the transition function is updated after every step (Line 12). New $Q$-value estimates are computed every time after an update of the transition function (Line 13). Note that we use a separate GP to estimate the transition function in each simulator.

One can use a number of termination conditions (Line $5$), \eg, maximum number of steps, changes in the value function or maximum number of switches. In our implementation, Algorithm~\ref{mbGP:MFRL} terminates if the change in new estimates of value functions in the real-world environment is no more than a certain threshold (ten percent) compared to the previous estimates.

The planner utilizes the knowledge of transitions from higher simulators (Lines 26--28) as well as lower simulators (Line 25) to encourage exploration in the current simulator. For every state action-action pair $(s, a)$, the planner looks for the maximum fidelity simulator in which a known estimate of transitions for  $(s, a)$ is available and uses them to plan in the current simulator. An estimate is termed known if the variance is below a threshold. If no such simulator is available, then it uses the $Q$-values learned in the previous simulator plus a fidelity parameter $\beta$. This parameter models the maximum possible difference between the optimal $Q$-values in consecutive simulators. \revone{The higher fidelity simulator values will always be trusted over the lower fidelity ones as long as we have low enough uncertainty in those estimates. We assume that, for two consecutive simulators, the maximum difference between the optimal action value of a state-action pair in $\Sigma_i$ and corresponding pair in $\Sigma_{i-1}$ is not more than $\beta_i$.} Note that one needs to apply a state-space discretization to plan the actions. However, the learned transition function is continuous.

\subsection{GPQ-MFRL Algorithm}
The agent learns optimal $Q$-values using GPs directly, instead of learning the model first.
The underlying assumption is that nearby state-action pairs will produce similar $Q$-values. This assumption can also be applied to problems where the states and actions are discrete but the transition function implies some sense of continuity. We choose the squared-exponential kernel because it models the spatial correlation we expect to see in a ground robot. However, any appropriate kernel can be used. We use a separate GP per simulator to estimate the $Q$-values using only data collected in that simulator.
\begin{algorithm}[htp]
\caption{GPQ-MFRL Algorithm}\label{mfGP:MFRL}
\begin{algorithmic}[1]
\Procedure{}{}
\BState \textbf{Input:} confidence parameters $\sigma_{th}$ and $\sigma^{sum}_{th}$; simulator chain $\langle \text{\boldmath$\Sigma$} 
,~\textnormal{fidelity parameters}~\text{\boldmath$\beta$},~\textnormal{state mappings}~\text{\boldmath$\rho$} \rangle$; $\mathcal{L}$.  
\BState \textbf{Initialize:} $\hat{Q}_i=\textnormal{initialize GP for}~i\in\lbrace 1, \ldots, d \rbrace$; state $s_0$ in simulator $\Sigma_1;~i~\gets~1;~change=\textsc{False}$. 
\BState \textbf{Initialize:} $t~\gets~0$; $\mathcal{D}_i~\gets~\lbrace\rbrace~\textnormal{for}~i\in\lbrace 1, \ldots, d \rbrace$. 
\BState \textbf{while} terminal condition is not met
\State  \ \  $a_t~\gets~\textnormal{\textsc{chooseaction}($s_t, i$)}$ 
\State \ \   \textbf{if} $\sigma_i(s_t, a_t)\leq \sigma_{th}$: $change=\textsc{True}$
\State \ \    \textbf{if} $\sigma\left(\rho_i(s_t), a_t\right) > \sigma_{th}$ and $change$ and $i>1$
\State \ \ \ \ \ \ \ \hspace{.25in} $s_t~\gets~\rho_i(s_t),~i~\gets~{i-1}$, continue
\State \ \    $\langle r_t, s_{t+1} \rangle~\gets~\textnormal{execute action}~a_t~\textnormal{in}~\Sigma_i$
\State \ \ $\textnormal{append}~\langle s_t, a_t, s_{t+1}, r_t \rangle~\textnormal{ to $\mathcal{D}_i$}$
\State \ \    $\mathcal{Y}_i~\gets~\lbrace\rbrace$ 
\State \ \ \textbf{for} $\langle s_t, a_t, s_{t+1}, r_t \rangle \in~\mathcal{D}_i$ //batch training// 
\State \ \ \ \ \ \ \ \hspace{.25in} $y_t~\gets~r_t+\gamma \textnormal{max}_a \hat{Q}_i\left(s_{t+1}, a\right)$
\State \ \  \ \ \ \ \hspace{.25in} $\textnormal{append}~\langle s_t, a_t, y_t \rangle~\textnormal{to}~\mathcal{Y}_i$  
\State \ \ $\hat{Q}_i~\gets~\textnormal{update $\textnormal{GP}_i$ using}~\mathcal{Y}_i$ 
\State \ \ \  \textbf{if} $\sum_{j=t}^{j=t-\mathcal{L}}\sigma_i(s_j, a_j) \leq \sigma^{sum}_{th}~\textnormal{and}~t>\mathcal{L}~\textnormal{and}~i<d$ 
\State  \ \ \ \ \ \ \ \hspace{.25in} $s_t~\gets~\rho_{i+1}^{-1}\left(s_t\right),~i~\gets~i+1$
\State  \ \ \ \ \ \ \ \hspace{.25in} $change=\textsc{False}$
\BState \textbf{end procedure}
\EndProcedure \\
\Procedure{chooseAction$(s, i)$}{}
\BState \textbf{for} $a\in\mathcal{A}(s)$ 
\State  $Q(s, a) = \hat{Q}_{i-1}\left(\rho_i(s), a\right) + \beta_i$ \State \textbf{for} $k\in\lbrace i, \ldots, d\rbrace$ 
\State \ \ \ \ \ \ $s_k = \rho^{-1}_{k}\ldots\rho^{-1}_{i+2}\rho^{-1}_{i+1}(s)$ 
\State \ \ \ \ \ \ \textbf{if} $\sigma_k(s_k, a)\leq\sigma^{th}$: $Q(s, a) = \hat{Q}_k(s_k, a)$ 
\BState return $\argmax_a Q(s, a)$
\BState \textbf{end procedure}
\EndProcedure
\end{algorithmic}
\end{algorithm}
Algorithm~\ref{mfGP:MFRL} gives the details of the proposed framework.  
GPQ-MFRL continues to collect samples in the same simulator until the agent is confident about its optimal actions. If the running sum of the variances is below a threshold (Line 17), this suggests that the robot has found a good policy with high confidence in the current simulator and it must advance to the next one (Line 18).

GPQ-MFRL uses similar thresholds $(\sigma_{th}~\textnormal{and}~\sigma_{th}^{sum})$ as GP-VI-MFRL to decide when to switch to a lower or higher fidelity simulator. GP-VI-MFRL checks if the agent has a sufficiently accurate estimate of the transition function in the previous simulator while GPQ-MFRL checks if the agent has a sufficiently accurate estimate of optimal $Q$-values in the previous simulator (Line $8$). Lines 10--15 describe the main body of the algorithm where the agent records the observed transitions in $\mathcal{D}_i$. We update target values (Line 14) for every transition as more data gets collected in $\mathcal{D}_i$ (Line $13$). The GP model is updated after every step (Line 16).

The agent utilizes the experiences collected in higher simulators (Lines 25--27) to choose the optimal action in the current simulator (Line $6$). Specifically, it checks for the maximum fidelity simulator in which the posterior variance for $(s, a)$ is less than a threshold $\sigma_{th}$. If one exists, it utilizes the $Q$-values from the highest known simulator to choose next action in the current simulator. If no such higher simulator exists, the $Q$-values from the previous simulator (Line 24) are considered to choose the next action in the current simulator with an additive fidelity parameter $\beta$.

GPQ-MFRL performs a batch retraining every time the robot collects new sample in a simulator (Lines 13--15). During the batch retraining, the algorithm updates the target values in previously collected training data using the knowledge gained by collecting new samples. Then these updated target values are used to predict the $Q$-values using GPs (Line 16). As the amount of data grows, updating the GP can become computationally expensive. However, we can prune dataset using sparse GP techniques~\cite{quinonero2005unifying}. 
It is non-trivial to choose values for confidence bounds but for the current experiments we chose the $\sigma_{th}^{sum}$ to be ten percent of the maximum $Q$-value possible and $\sigma_{th}$ to be one fifth of $\sigma_{th}^{sum}$.  

\section{Results} \label{section:results}
We use two environments to simulate GP-VI-MFRL and three environments for GPQ-MFRL. For GP-VI-MFRL, the goal is learning to navigate from one point to another while avoiding the obstacles. $\Sigma_1$ is a $21\times 21$ grid-world with a point robot whereas $\Sigma_2$ is Gazebo (discretized in $21\times 21$ grid) which simulates the kinematics and dynamics of a quadrotor operating in 3D. For GPQ-MFRL, the goal is to learn avoiding the obstacles while navigating through the environment. $\Sigma_1$ is Python-based simulator Pygame, $\Sigma_2$ is a Gazebo environment, and $\Sigma_3$ is the real world. \revone{We further use sparse GPs to speed up the computations required to perform GP inference. We report the improvements in computational time as well as a direct comparison between GP-VI-MFRL and GPQ-MFRL on an obstacle avoidance task.}

\subsection{GP-VI-MFRL Algorithm}
\begin{figure}[htp]
	\centering
	\includegraphics[width=1\columnwidth]{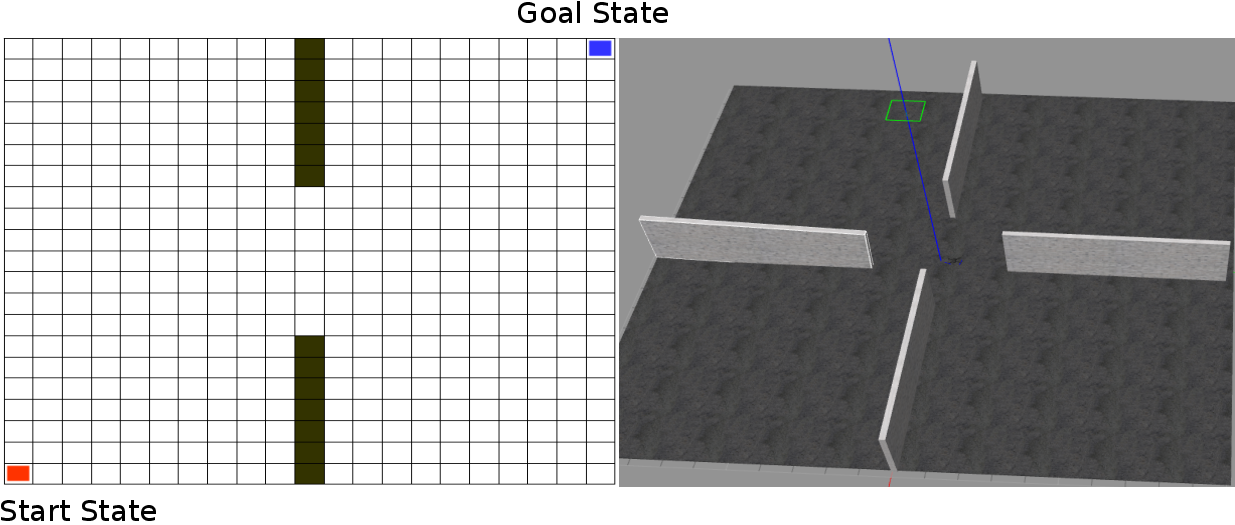}
	\caption{The environment setup for a multi-fidelity simulator chain. The grid-world simulator $(\Sigma_1)$ has two walls whereas the Gazebo simulator $(\Sigma_2)$ has four walls as shown.}
   \label{fig:gp_mfrl_setup}
\end{figure}

The task of the robot is to navigate from the start state to goal state. The start and goal states and the obstacles for the environment used are shown in Figure~\ref{fig:gp_mfrl_setup}. The state of the robot is given by its $X$ and $Y$ coordinates whereas the action is a 2D velocity vector. Both simulators have the same state-space, therefore, $\rho_i$ is an identity mapping. The robot gets a reward of zero for all transitions except when it hits the obstacles in which case it gets a reward of -50 and a reward of 100 for landing in the goal state.

Since the state space $\mathcal{S}\in\mathbb{R}^2$ and the action space (velocity)  $\mathcal{A}\in\mathbb{R}^2$, the true transition function is $\mathbb{R}^4\rightarrow\mathbb{R}^2$. However, generally GP regression allows for single-dimensional outputs only. Therefore, we assume independence between the two output dimensions and learn two components (along $X$ and $Y$) of the transition functions separately, $x_{i+1}=f_x(x_i,y_i,a_x)$ and $y_{i+1}=f_y(x_i,y_i,a_y)$, where $(x_i,y_i)$ and $(x_{i+1},y_{i+1})$ are the current and next states of the robot, and $(a_x,a_y)$ is the velocity input. The GP prediction is used to determine the transitions, $(x_i,y_i,a_x)$ $\rightarrow$ $x_{i+1}$ and $(x_i,y_i,a_y)$ $\rightarrow$ $y_{i+1}$ where $(x_{i+1},y_{i+1})$ is the predicted next state with variances $\sigma_x^2$ and $\sigma_y^2$ respectively. 

\begin{figure}[htp]
	\centering 
    \includegraphics[width=5cm]{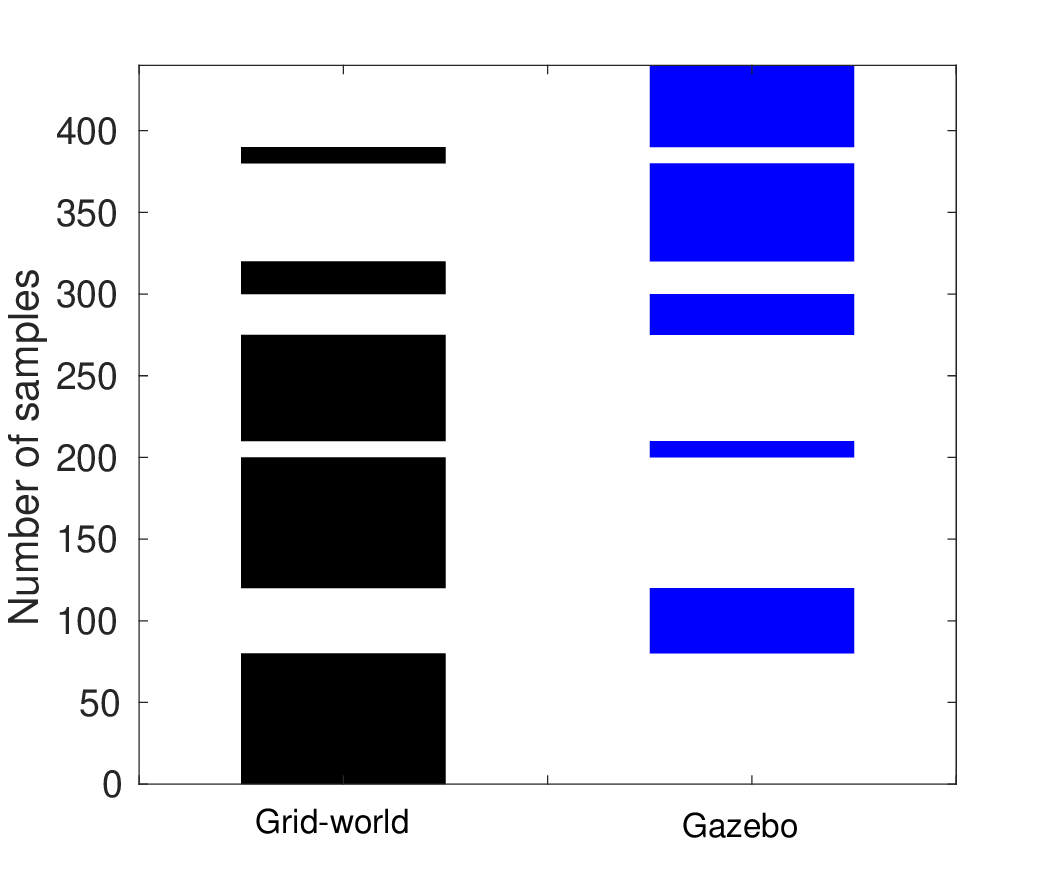}
	\caption{ The figure represents the samples collected at each level of the simulator for a $21 \times 21$ grid in a grid-world and Gazebo environments. $\sigma_{th}^{sum}$ and $\sigma_{th}$ were kept 0.4 and $0.1$ respectively.}
   \label{fig:epoch_samples}
\end{figure}

Figure~\ref{fig:epoch_samples} shows the switching between the simulators for one run of the GP-VI-MFRL algorithm on the simulators shown in Figure~\ref{fig:gp_mfrl_setup}. Unlike unidirectional transfer learning algorithms, GP-VI-MFRL agent switches back-and-forth in simulators collecting most of the samples in the first simulator initially. Eventually, the robot starts to collect more samples in the higher fidelity simulator. This is the case when the algorithm is near convergence and has accurate estimates for transitions in lower fidelity simulator as well. Next, we study the effect of the parameters used in GP-VI-MFRL and the fidelity of the simulators on the number of samples until convergence.

\subsubsection{Variance in learned transition function}
\begin{figure}
	\centering 
    \subfigure[Initialization]{\includegraphics[width=0.45\columnwidth]{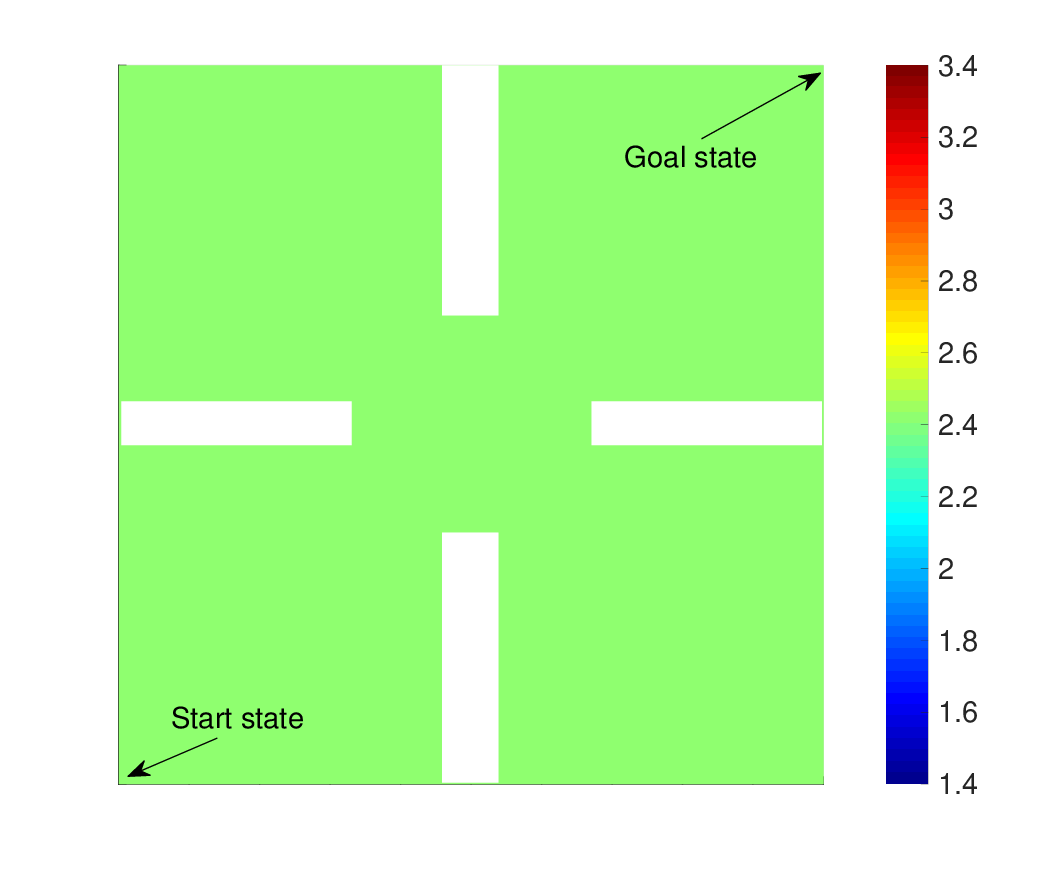}}
    \subfigure[After convergence]{\includegraphics[width=0.45\columnwidth]{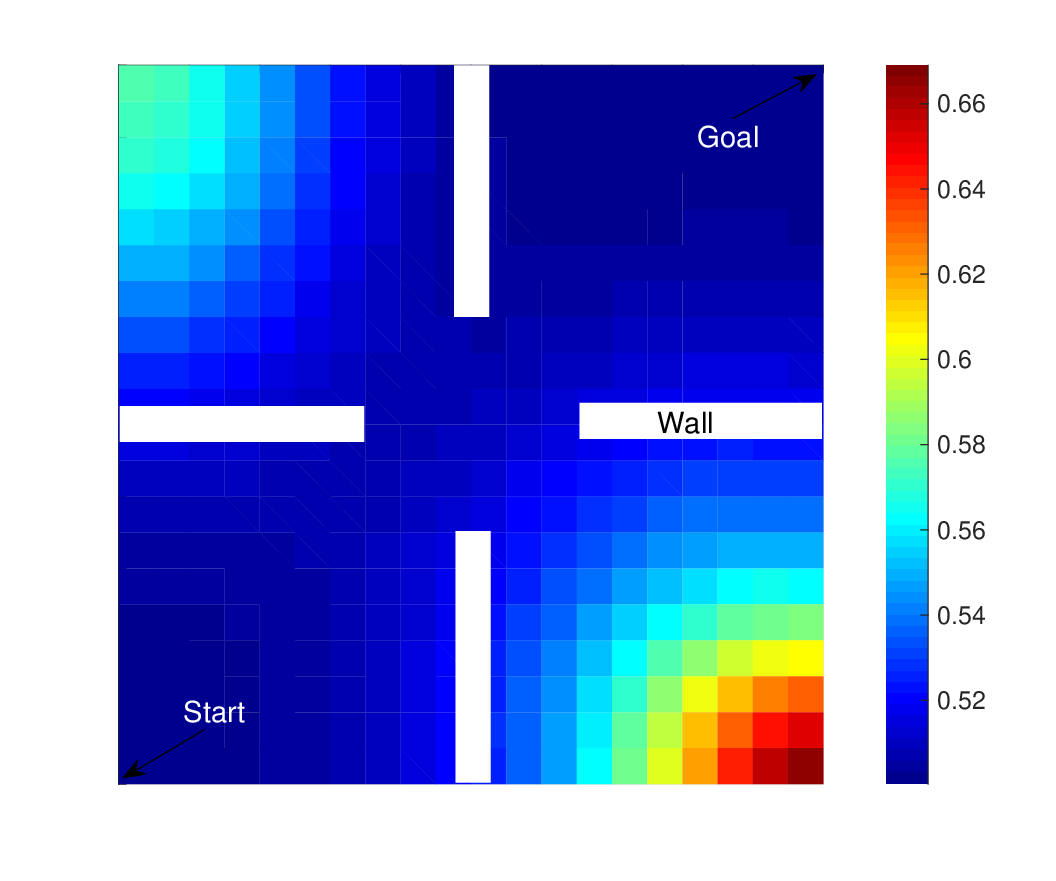}}
	\caption{Variance plot for Gazebo simulator after transition function initialization and after the algorithm has converged. Colored regions show the respective $\sqrt{\sigma_x^2 + \sigma_y^2}$ values for the optimal action returned by the planner in each state.}
   \label{fig:heatmap2}
\end{figure}

To demonstrate how the variance of the predicted transition function varies from the beginning of the experiment to convergence, we plot ``heatmaps'' of the posterior variance for Gazebo environment transitions. The GP prediction for a state-action pair gives the variance, $\sigma_x^2$ and $\sigma_y^2$, respectively for the predicted state. After convergence (Figure~\ref{fig:heatmap2}), the variance along the optimal (\ie, likely) path is low whereas the variance for states unlikely to be on the optimal path from start to goal remains high since those states are explored less often in Gazebo environment. Hence utilizing the experience from lower fidelity simulator results in more directed exploration of the higher fidelity simulators.

\subsubsection{Effect of fidelity on the number of samples}
\begin{figure}
	\centering 
    \subfigure[The ratio of samples collected in Gazebo and total samples (y-axis) as a function of the fidelity of grid-world. We lower the fidelity of grid-world by increasing the variance of the simulated transition function.\label{fig:gp_mfrl_ratio}]{\includegraphics[width=0.45\columnwidth]{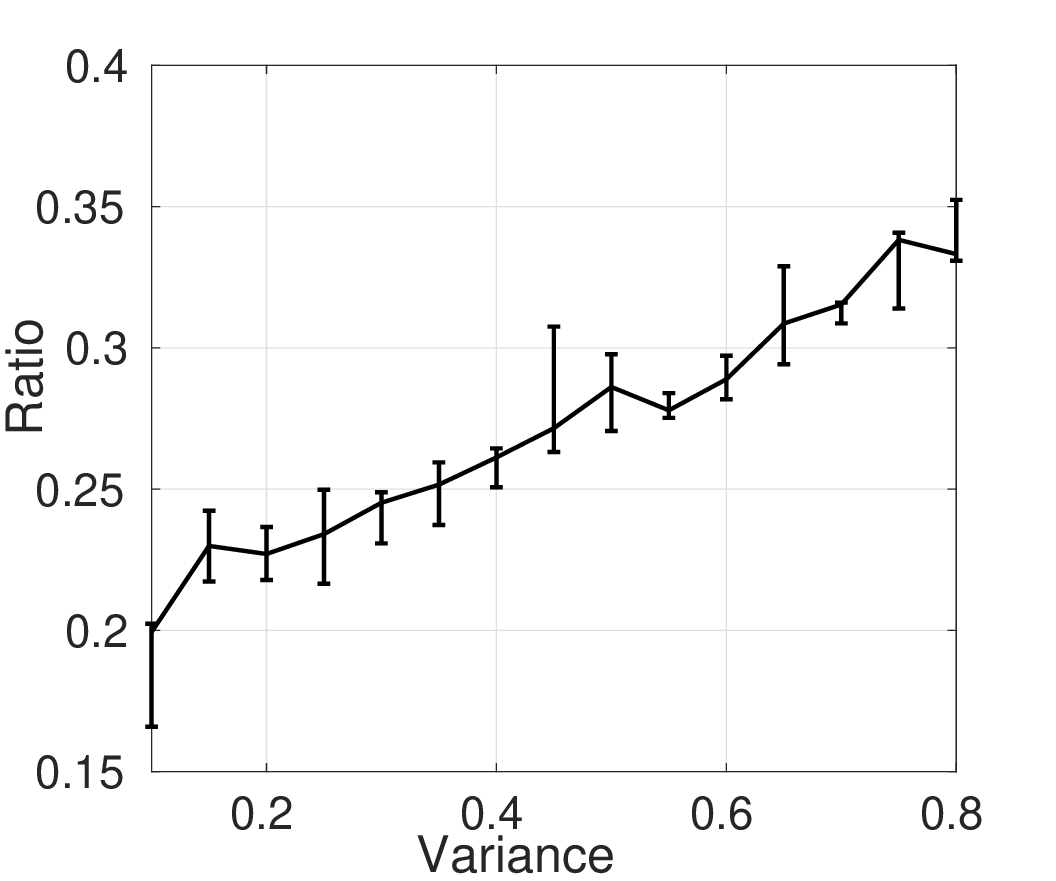}}
  \subfigure[The number of samples collected (y-axis) in Gazebo increases more rapidly (as demonstrated by diminishing vertical separation between the two plots) than samples collected in grid-world. \label{fig:gp_mfrl_samples}]{\includegraphics[width=0.45\columnwidth]{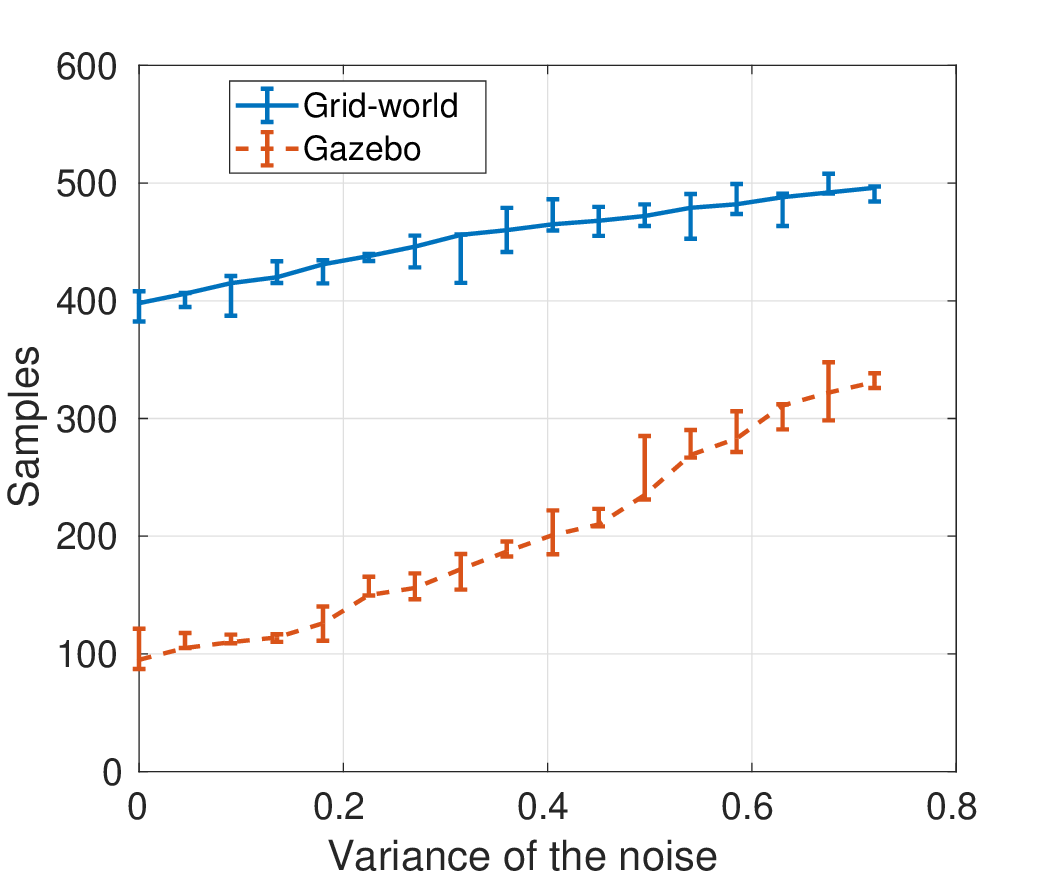}}
	\caption{As we lower the fidelity of grid-world by adding more noise in grid-world transitions, the agent tends to spend more time in Gazebo. The plots show the average and min-max error bars of $5$ trials.}
   \label{fig:gp_mfrl_}
\end{figure}

Next, we study the effect of varying the fidelity on the total number of samples and the fraction of the samples collected in the Gazebo simulator. Our hypothesis is that as the fidelity of the first simulator decreases, the agent will need more samples in Gazebo. In order to validate this hypothesis, we varied the noise added to simulate the transitions in the grid-world. The transition model in Gazebo remains the same. The total number of samples collected increases as we increase the noise in grid-world (Figure \ref{fig:gp_mfrl_samples}). As we increase the noise in grid-world, the agent learns less accurate transition function leading to more samples collected in Gazebo. Not only does the agent need more samples, the ratio of the samples collected in Gazebo to the total number of samples also increases (Figure \ref{fig:gp_mfrl_ratio}). 

\subsubsection{Effect of the confidence parameters}
\begin{figure}[ht]
	\centering
	\includegraphics[width=0.8\columnwidth]{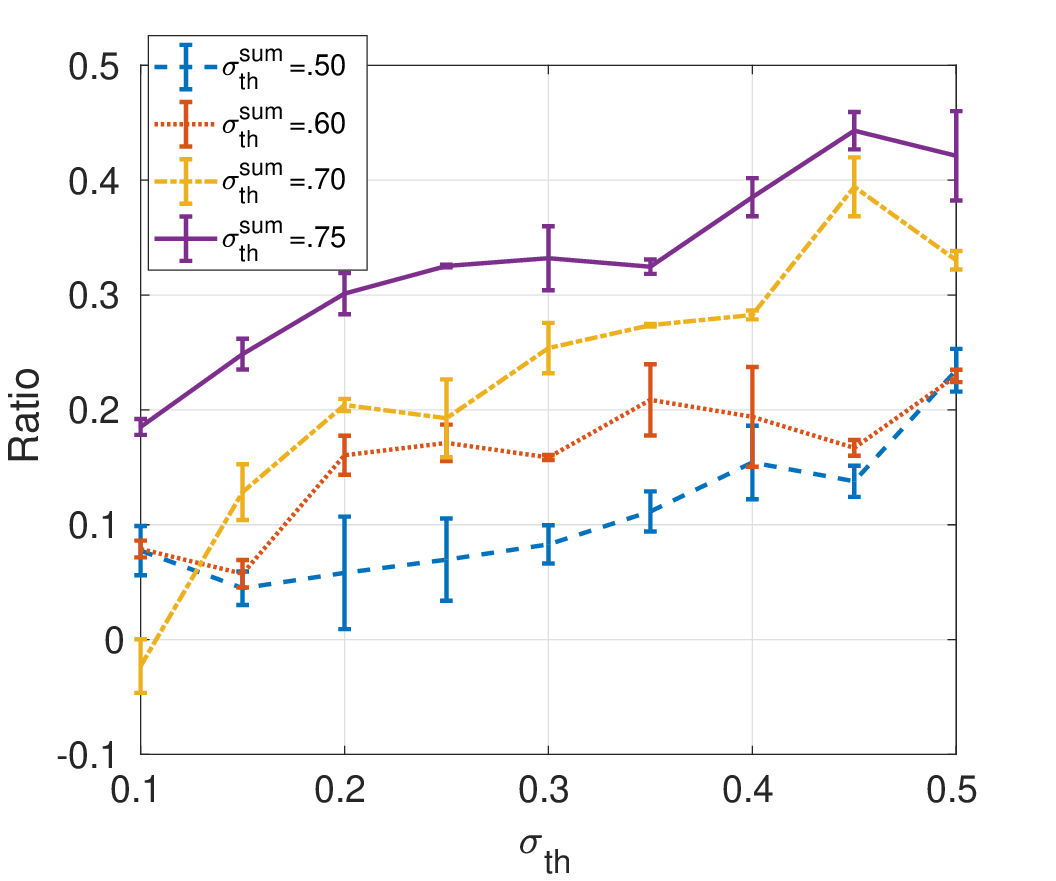} 
	\caption{The ratio of samples collected in Gazebo to the total samples as a function of confidence parameter $\sigma_{th}$ for four different values of $\sigma_{th}^{sum}$. The figure shows the average and standard deviation of 5 trials.}
   \label{fig:threshold}
\end{figure}

GP-VI-MFRL algorithm uses two confidence parameters, $\sigma_{th}$ and $\sigma_{th}^{sum}$, which quantify the variances in the transition function to switch to a lower and higher simulator, respectively. Figure~\ref{fig:threshold} shows the effect of varying the two parameters on the ratio of the number of samples collected in Gazebo simulator to the total number of samples. Smaller $\sigma_{th}$ and $\sigma_{th}^{sum}$ results in the agent collecting more samples in the lower fidelity simulator and may result in slow convergence. Depending on the user preference, one can choose the values of confidence bounds from the Figure~\ref{fig:threshold}.

\subsubsection{Comparison with RMax MFRL}
\begin{figure}[ht]
	\centering 
    \includegraphics[width=0.8\columnwidth]{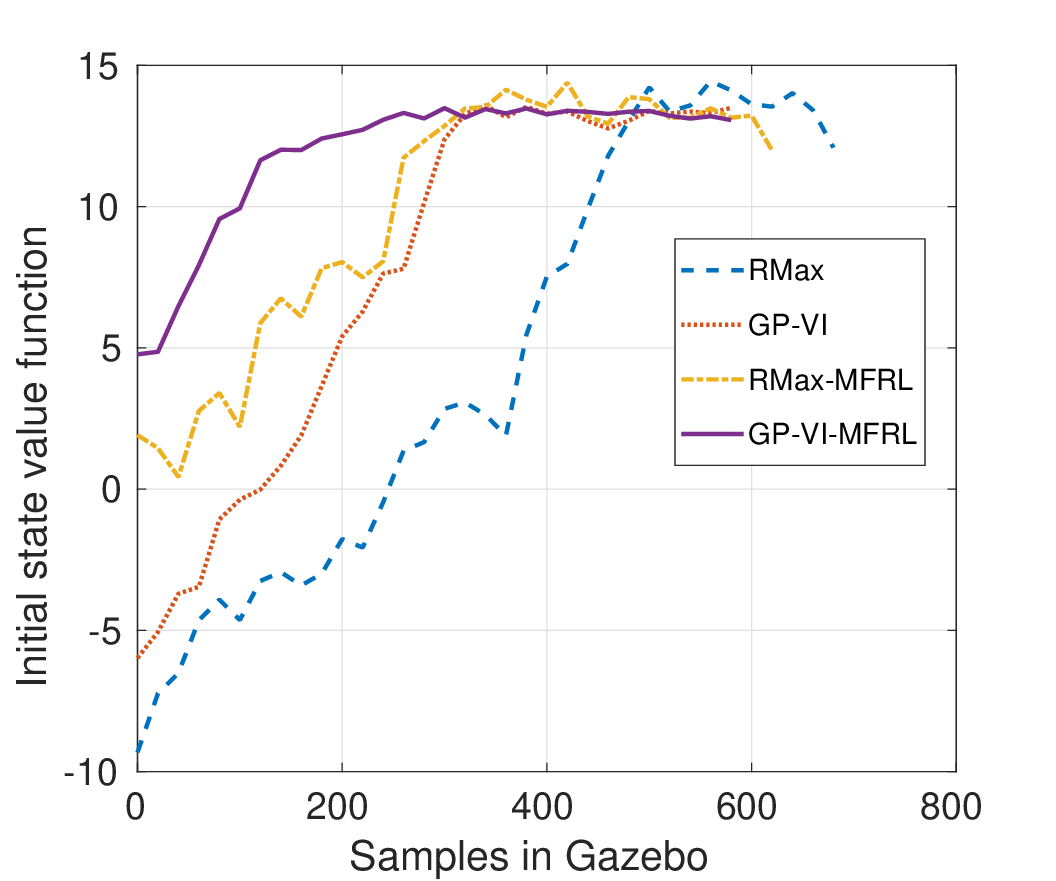}
	\caption{Comparison of GP-VI-MFRL with three baseline strategies. The $Y$-axis shows the value function for the initial state $\left(V(s_0)\right)$ in Gazebo as a function of the number of samples collected in Gazebo. Value function estimation for GP-VI-MFRL converges fastest.}
   \label{fig:value_function}
\end{figure}

Figure \ref{fig:value_function} compares GP-VI-MFRL with three other baseline algorithms, 
\begin{enumerate}
\item RMax algorithm running only in Gazebo without grid-world (RMax),
\item GP-MFRL algorithm only in Gazebo with no grid-world present (GP-VI) and
\item Original MFRL algorithm~\cite{cutler2014reinforcement} (RMax-MFRL).
\end{enumerate}

Specifically, we plot the value of the initial state, $V(s_0)$, as a function of the number of samples in Gazebo, \ie, $\Sigma_2$. We observe that GP-VI-MFRL uses fewer samples in Gazebo to converge to the optimal value than the other methods.

GP-VI-MFRL performs a GP update at each time step. GP update grows cubically with number of training samples which will make GP-VI-MFRL computationally infeasible beyond a certain number of training samples. However this issue can be addressed by using appropriate active learning strategies which select a subset of samples to retain thereby keeping the size of the dataset constant. The total computational time for GP-VI-MFRL to perform GP updates on collected samples accounts for approximately 10 minutes. 

\subsection{GPQ-MFRL Algorithm}
\begin{figure}[ht]
	\centering 
    \includegraphics[width=\columnwidth]{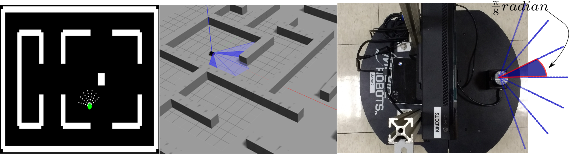}
	\caption{We use the Python-based simulator Pygame as $\Sigma_1$, Gazebo as $\Sigma_2$ and Pioneer P3-DX robot in real-world as $\Sigma_3$.}
   \label{setup:gpq}
\end{figure}

We use three environments (Figure~\ref{setup:gpq}) to demonstrate the GPQ-MFRL algorithm. The task for the robot is to navigate through a given environment without crashing into the obstacles, assuming the robot has no prior information about the environments. There is no goal state.

The robot has a laser sensor that gives distances from obstacles along seven equally spaced directions. The angle between two consecutive measurement directions was set to be $\frac{\pi}{8}$ radians. The actual robot has a Hokuyo laser sensor that operates in the same configuration. Distance measurements along the seven directions serve as the state in the environments. Therefore we have a seven-dimensional continuous state space: $\mathcal{S}\in (0, 5]^7$. The linear speed of the robot was held constant at $0.2$ m/sec. The robot can choose its angular velocity from nineteen possible options: $\lbrace-\frac{\pi}{9}, -\frac{\pi}{8}, \ldots, \frac{\pi}{9} \rbrace$. The reward in each state was set to be the sum of laser readings from seven directions except when the robot hits the obstacle. In case of a collision, it gets a reward of -50.

We train the GP regression, $Q(\textbf{s}, a):\mathbb{R}^8 \rightarrow \mathbb{R}$. Hyperparameters of the squared-exponential kernel were calculated off-line by minimizing the negative log marginal likelihood of 2000 training points which were collected by letting the robot run in the real world directly. The parameter values for experiments in this section are given in Table~\ref{tab:par}.

\begin{table}[ht]
\caption{Parameters used in GPQ-MFRL} 
\centering 
\begin{tabular*}{\columnwidth}{c @{\extracolsep{\fill}} l c r}
\hline\hline 
 Description &  Type &\multicolumn{1}{c}{Value}
\\ 
\hline 
  &$\sigma$ &  102.74 \ \ \ \ \ \ \ \ \ \ \ \\[-1ex]
\raisebox{1.5ex}{Hyperparameters}
&\textit{l} & [2.1, 5.1, 14, 6.2, 15, 2, 2, 1]  \\[1ex]
&$\omega^2$ & 20 \ \ \ \ \ \ \ \ \ \ \ \ \ \  \\[0.2ex]
\hline
 &$\sigma_{th}^{sum}$ & 60 \ \ \ \ \ \ \ \ \ \ \ \ \ \   \\[-1ex]
\raisebox{1.5ex}{Confidence parameters} 
& $\sigma_{th}$ &15 \ \ \ \ \ \ \ \ \ \ \ \ \ \  \\[0.2ex]
\hline
\raisebox{.5ex}{Algorithm} 
& $\mathcal{L}$ & 5 \ \ \ \ \ \ \ \ \ \ \ \ \ \ \ \\
\hline 
\end{tabular*}
\label{tab:par}
\end{table}

\subsubsection{Average Cumulative Reward in Real-World}
\begin{figure}[htp]
	\centering
	\includegraphics[width=\columnwidth]{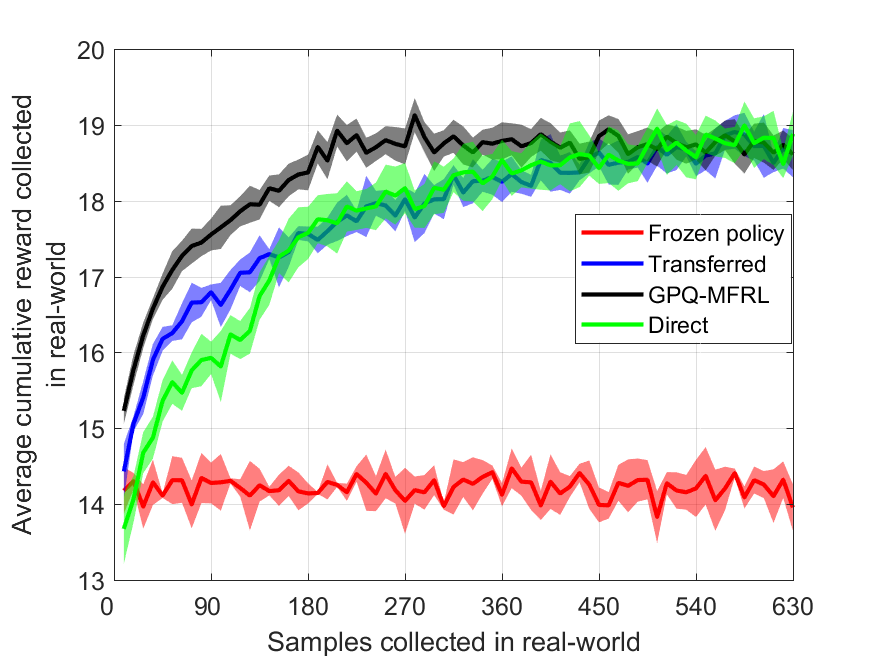}
	\caption{Average cumulative reward collected by the Pioneer robot in real-world environment as a function of the samples collected in the real world.  The plot shows the average and standard deviation of $5$ trials.}
   \label{fig:avg:reward}
\end{figure}

In Figure~\ref{fig:avg:reward}, we compare GPQ-MFRL algorithm with three other baseline strategies by plotting the average cumulative reward collected by the robot as a function of samples collected in the real world. Three baseline strategies are, 
\begin{enumerate}
\item Directly collecting samples in real world without the simulators (Direct),
\item Collect hundred samples in one simulator and transfer the policy to the Pioneer robot with no further learning in the real world (Frozen Policy) and
\item Collect hundred samples in one simulator and transfer the policy to the robot while continuing to learn in the real world (Transferred Policy).
\end{enumerate}    
We observe that the Direct \revone{policy} performs worst in the beginning. It can be attributed to the fact that the robot started to learn from scratch. The Frozen policy starts \revone{better} since it \revone{has already} learned a policy in the simulator. However, it \revone{tends} to a lower value of average cumulative reward which suggests that the optimal policy learned in the simulator is not the optimal policy in the real world. Although, the Transferred policy seems to perform better at the beginning than the Frozen policy, it is difficult to dictate if it will always be the case. The Direct policy has a large performance variance in the beginning. GPQ-MFRL outperforms the other strategies right from the beginning. \revone {We attribute this to the fact that} the GPQ-MFRL collects more samples from the simulator in the beginning and hence starts \revone{better} right from the \revone{start}. If one would allow the Transferred Policy and the Frozen Policy to collect more samples from the simulator they might have performed the same as the GPQ-MFRL. However, deciding how many samples one should allow is a non-trivial task and problem-specific. GPQ-MFRL can decide the number of samples in each simulator by itself without the need for human intervention.

\subsubsection{Policy Variation with Time} 
\begin{figure}
	\centering 
    \subfigure[Sum of absolute change in value functions]{\includegraphics[width=4.1cm]{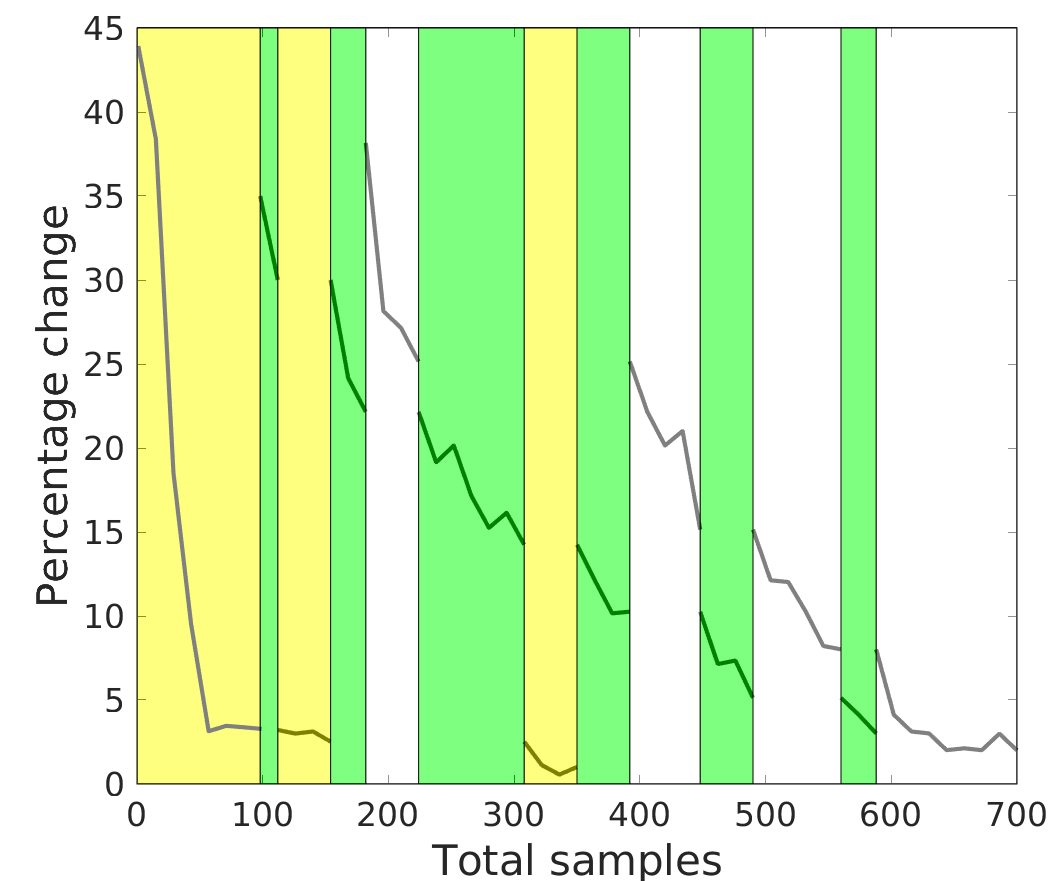}}
    \subfigure[Average variances in value function estimations]{\includegraphics[width=4.2cm]{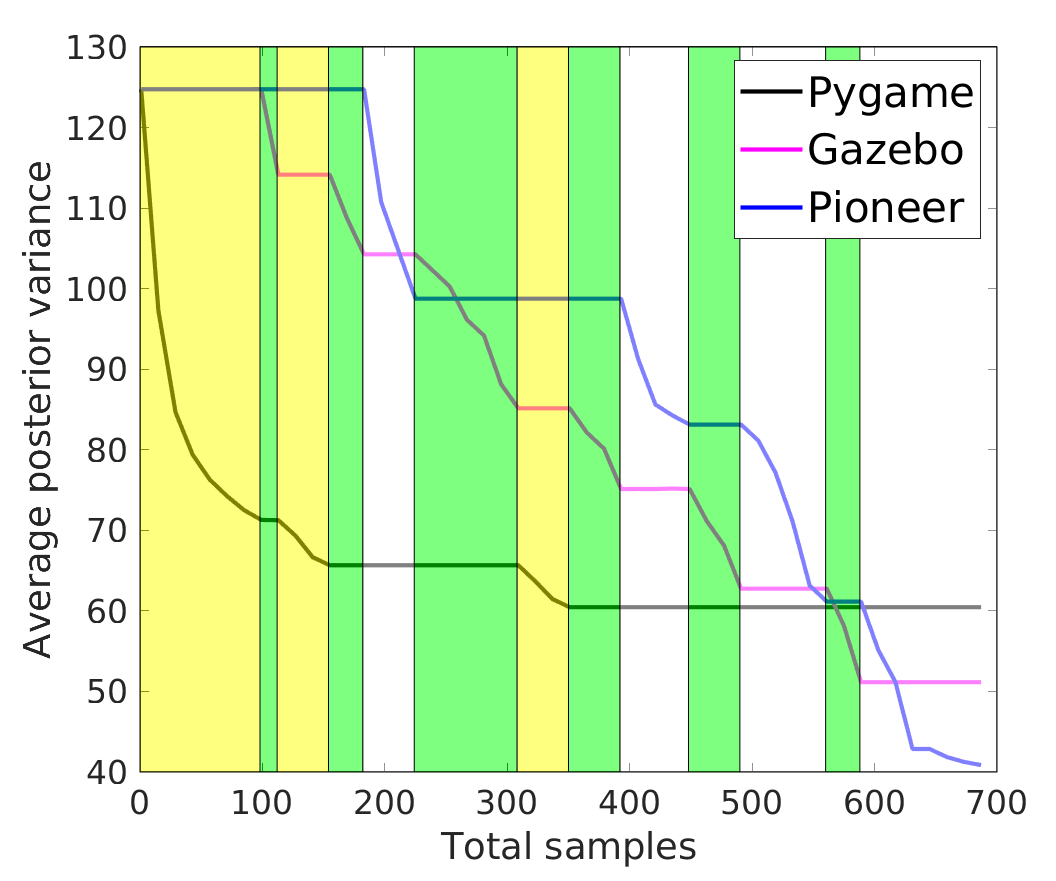}}
	\caption{ Yellow, green and white regions correspond to the samples collected in the Pygame, Gazebo and real-world environments respectively. Plots are for state set $\lbrace 1, 3, 5\rbrace^7$.}
   \label{fig:states_float}
\end{figure}

Figure~\ref{fig:states_float} shows the absolute percentage change in the sum of the value functions with respect to last estimated sum of value functions and average predictive variance for states $\lbrace 1, 3, 5\rbrace^7$ in all three simulators. Observe that initially most of the samples are collected in the simulator, whereas over time the samples are collected mostly in the real-world. The simulators help the robot to make its value estimates converge quickly as observed by a sharp dip in the first white region. Note that GP updates for $i^{th}$ simulator $(\hat{Q}_i)$ are made only when the robot is running in $i^{th}$ simulator.     

\revone{
\subsubsection{Higher-dimensional spaces and sparse GPs}
One of the limitations of GPs is their computational complexity which grows cubically with the number of training samples. However, we use sparse approximations to address this limitation. We also increase the dimensionality of the state-space to verify if the proposed algorithms scale to higher dimensions. Specifically, we increase the number of laser readings to 180 equally spaced directions. Therefore, GP regression is used to estimate $Q(\textbf{s},a): \mathbb{R}^{181}\rightarrow\mathbb{R}$. }

\revone{There are several methods for sparse GP approximations. We use the technique from \cite{quinonero2005unifying}that finds a possibly smaller set of points (called inducing points) which fit the data best. The GP inference is conditioned on the smaller set of inducing points rather than the full set of training samples. Finding inducing points is closely related to finding low-rank approximations of the full GP covariance matrix. The inducing points may or may not belong to the actual training data.  }

\begin{figure}
	\centering 
    \subfigure[Inducing points set to $5 \%$ of the training sample size]{\includegraphics[width=4.1cm]{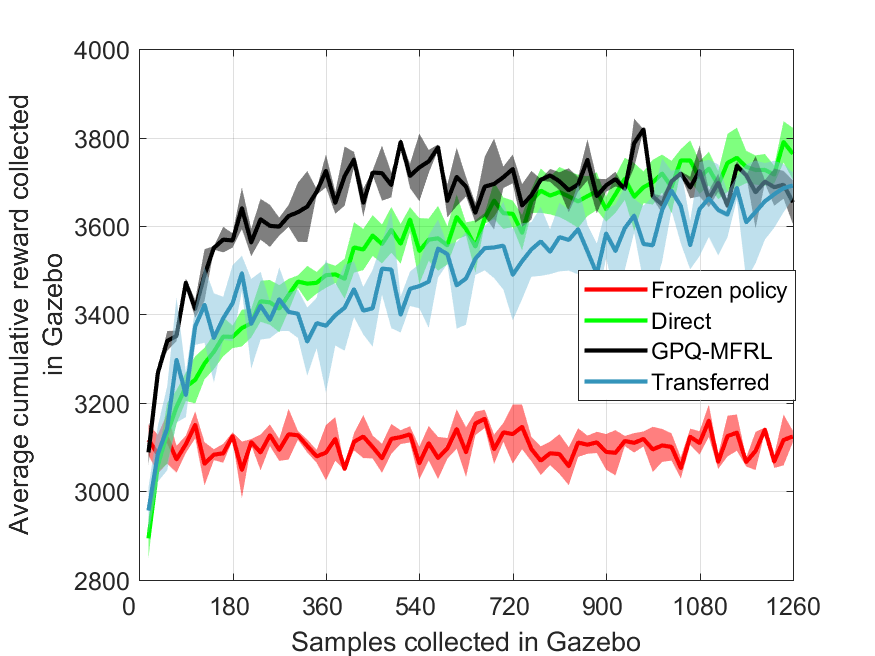}\label{fig:sparseGP5}}
    \subfigure[Inducing points set to $15 \%$ of the training sample size]{\includegraphics[width=4.2cm]{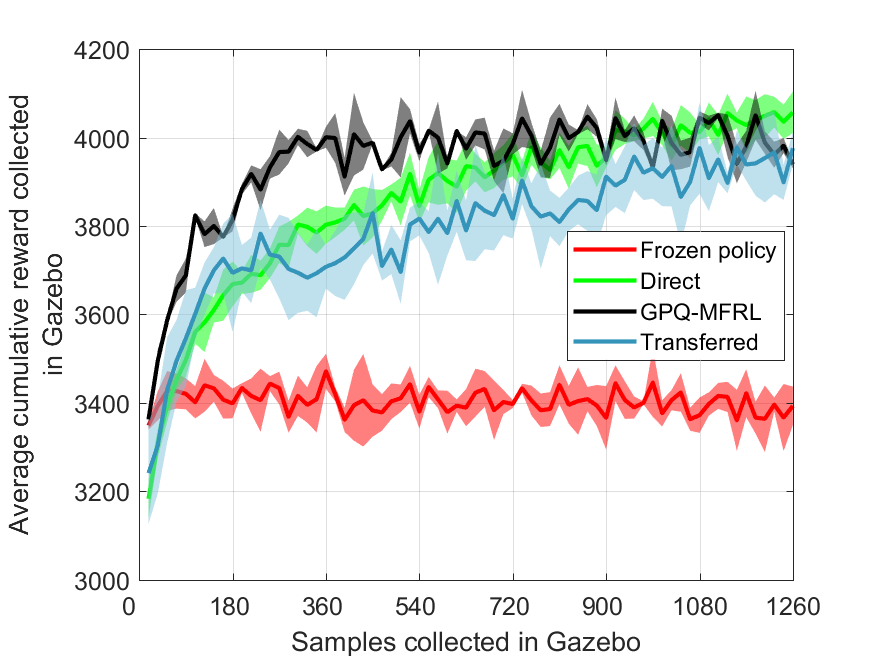}}
    \subfigure[Inducing points set to $25 \%$ of the training sample size]{\includegraphics[width=4.2cm]{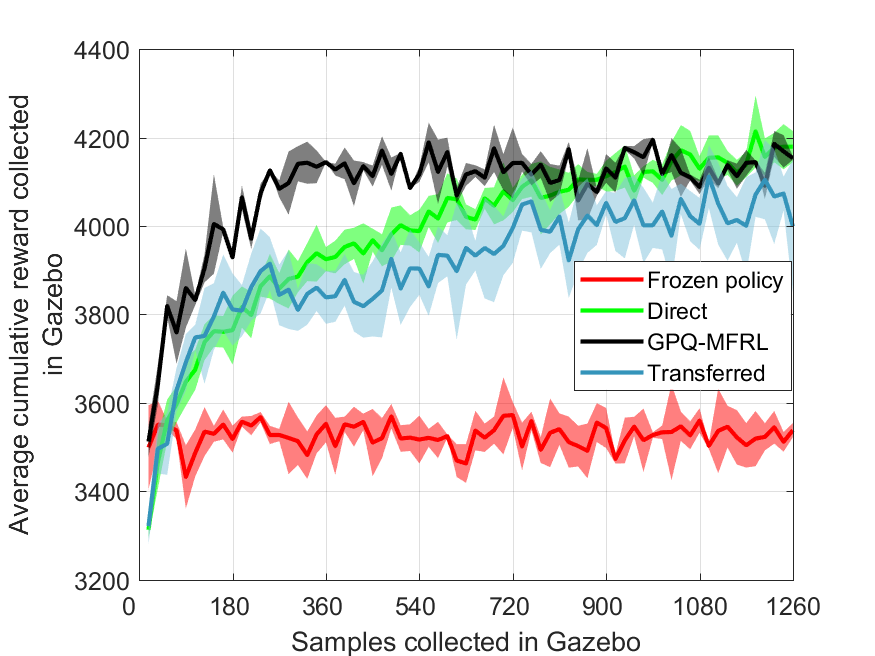}}
    \subfigure[Full GP inference]{\includegraphics[width=4.2cm]{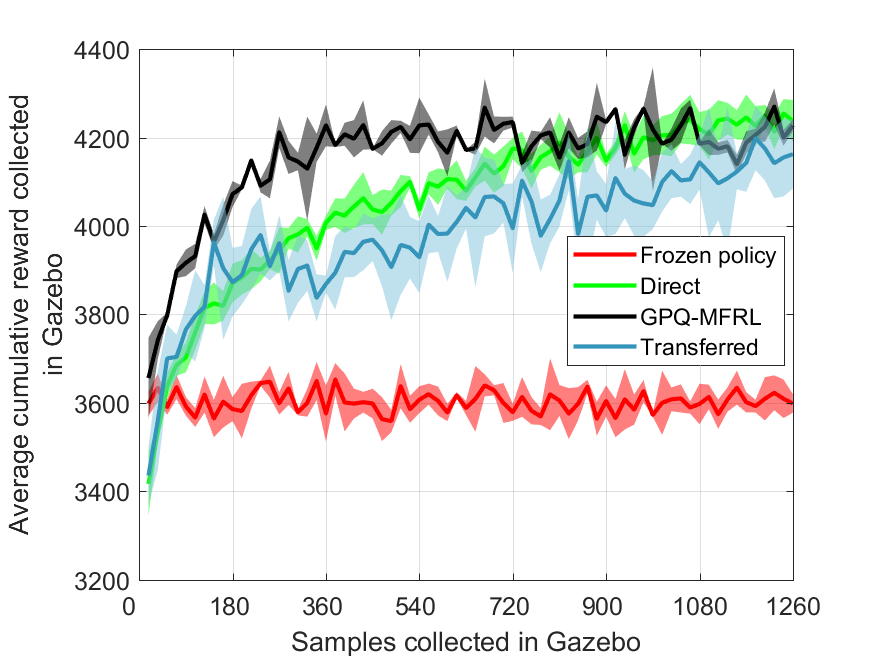}\label{fig:fullGP}}
	\caption{Average cumulative reward collected by the robot in Gazebo environment as a function of the samples collected in Gazebo for different percentage of inducing points. The plots show the average and standard deviation of $5$ trials.}
   \label{fig:avg_Cum_sparse}
\end{figure}

\begin{figure}
	\centering 
\includegraphics[width=\columnwidth]{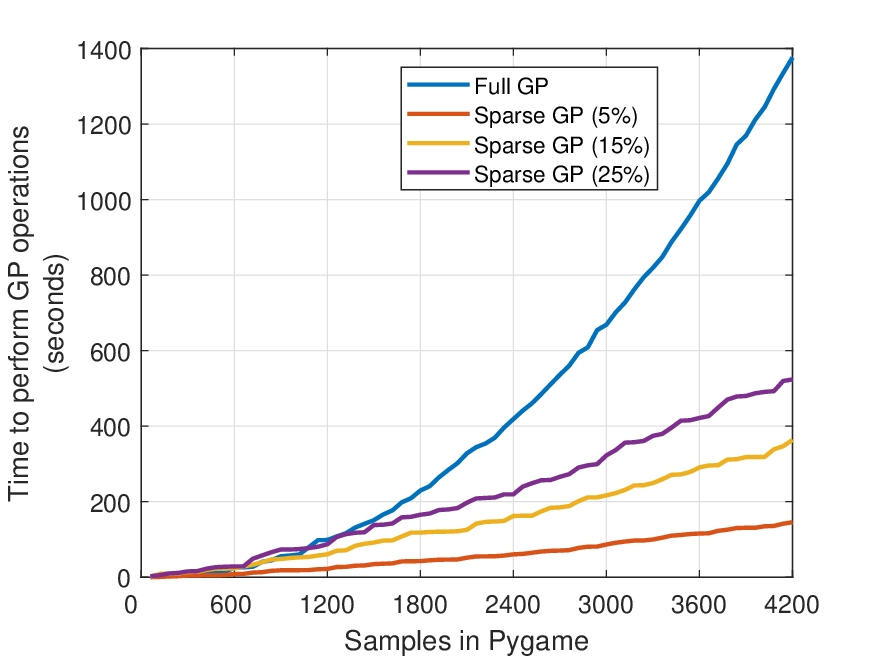}
	\caption{The wall clock time required to perform all GP related operations in Pygame for various degrees of GP sparse approximations in Pygame. The plot shows the mean time over five trials for each case.\label{fig:runPygame}}
   \label{fig:runningTime}
\end{figure}

\revone{We did several experiments with GPQ-MFRL to study the performance of the algorithm for number of inducing points. We use Pyro~\cite{bingham2018pyro} to implement sparse GPs. Figure~\ref{fig:avg_Cum_sparse} shows the average cumulative reward collected by the robot in the Gazebo environment when GP inference is done with the number of inducing points set to $5\%$, $15\%$, $25\%$ of the total training samples and using all the training samples. We observe a significant increase in cumulative reward collected going from $5\%$ to $15\%$ but not much from $15\%$ to $25\%$ (y-axes in Figure~\ref{fig:avg_Cum_sparse}). }
\revone{A plot of the wall clock times to perform GP inference in Pygame is shown in Figure~\ref{fig:runningTime}. The wall time to perform GP inference in Gazebo exhibits a similar trend which we omit for the sake of brevity. The wall clock time includes the time to perform all GP operations including the time to update the hyperparameters as well as finding the inducing points of both GPs. We update these after every ten new training samples in an individual simulator. The experiments were run on a machine running Ubuntu 16.04 with Intel(R) Core(TM) i7-5600U CPU @ 2.60 GHz, Intel HD Graphics 5500 and 16 GB RAM.}

\revone{The results suggests that a small fraction of inducing points are sufficient and yields diminishing marginal gains in the performance when the number of inducing points increases. Doing inference on inducing points with $25\%$ of the training data performs almost as good as doing full GP regression, in terms of the reward collected by the learned policy (Figure~\ref{fig:fullGP}) but is significantly faster. }

\begin{figure}[ht]
	\centering 
    \includegraphics[width=\columnwidth]{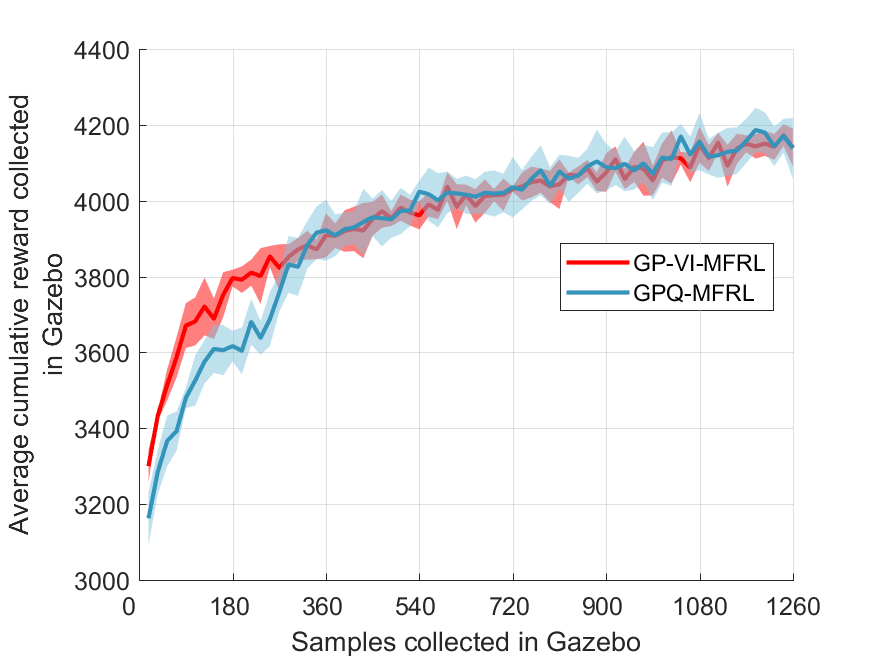}
	\caption{Average cumulative reward collected by the robot in Gazebo environment as a function of the samples collected in Gazebo for GP-VI-MFRL and GP-Q-MFRL. The plots show the average and standard deviation of $10$ trials.}
   \label{fig:QvsVI}
\end{figure}

\revone{\subsection{Comparison between GP-VI-MFRL and GPQ-MFRL}
We compare GP-VI-MFRL and GPQ-MFRL using average cumulative reward collected by the robot in Gazebo as the metric in the obstacle avoidance task. To do this comparison, we use full GP regression to perform the inference. The laser obtains distance measurements from seven equally spaced directions and we train seven independent GPs to learn the transition function in GP-VI-MFRL (one GP corresponding to each direction). A performance comparison is shown in Figure~\ref{fig:QvsVI}. Though both algorithms seem to perform the same asymptotically, GP-VI-MFRL performs slightly better than GPQ-MFRL in the beginning.}

\section{Discussion and Future Work} \label{sec:conc}
\revone{We present GP-based MFRL algorithms  that leverage multiple simulators with varying fidelity and costs to learn a policy in the real-world. We demonstrated empirically that the GP-based MFRL algorithms find the optimal policies using fewer samples than the baseline algorithms, including the original MFRL algorithm~\cite{cutler2014reinforcement}. 
The computational limitations of sparse GPs can be mitigated to an extent by the use of sparse GP approximations. We also provided a head-to-head comparison between the two algorithms presented here. GP-VI-MFRL (model-based version) performs better than GPQ-MFRL (model-free version) in the beginning. This is consistent with the outcomes for traditional RL techniques where model-based algorithms tend to perform better than model-free ones.}

\revone{An immediate future work is to compare the MFRL technique with sim2real approaches~\cite{chebotar2018closing}. Unlike sim2real, the presented MFRL techniques explicitly decide when to switch between simulators and use more than two levels of simulators. An interesting avenue of future work would be to combine the two ideas --- use MFRL to model the fact that some simulators are cheaper/faster to operate than others and use parameterized simulators to bring in domain adaptation/randomization for better generalization. Finally, in the current approach, data from different simulators are not combined when performing GP regression. One possibility of improvement is to use multi-task GPs that can simultaneously produce multiple outputs, one corresponding to each fidelity simulator.}

\revone{We can use multi-task GPs~\cite{bonilla2008multi} that can produce multiple outputs simultaneously, one corresponding to each simulator. An alternative is to use deep GPs to combine data from various fidelities as part of the same network. In both cases, the goal is to learn correlation between the values in different environments directly.}

\ifCLASSOPTIONcaptionsoff
  \newpage
\fi
\bibliographystyle{IEEEtran}
\bibliography{references}

\end{document}